\documentclass[11pt,a4paper]{article}
\usepackage[hyperref]{acl2019}
\usepackage{times}
\usepackage{microtype}
\usepackage{latexsym}
\usepackage{subcaption}
\usepackage{algorithm}
\usepackage{algorithmic}
\usepackage{amsmath}
\usepackage{hhline}
\usepackage{hyperref}
\usepackage{subcaption}
\usepackage{graphicx}
\usepackage{url}
\usepackage{bm}

\aclfinalcopy % Uncomment this line for the final submission
\title{Incremental Sense Weight Training for the\\Interpretation of Contextualized Word Embeddings}

\author{Xinyi Jiang \\
  EMORY University \\
  xinyi.jiang2@outlook.com \\\And
  Zhengzhe Yang \\
  EMORY University \\
  jackzz1997@outlook.com \\\And
    Jinho D. Choi \\
  EMORY University \\
  jinho.choi@emory.edu \\}

\date{}

\begin{document}
\maketitle

\begin{abstract}
We present a novel online algorithm that learns the essence of each dimension in word embeddings by minimizing the within-group distance of contextualized embedding groups. % to interpret the sense information 
Three state-of-the-art neural-based language models are used, Flair, ELMo, and BERT, to generate contextualized word embeddings such that different embeddings are generated for the same word type, which are grouped by their senses manually annotated in the SemCor dataset.
We hypothesize that not all dimensions are equally important for downstream tasks so that our algorithm can detect unessential dimensions and discard them without hurting the performance.
To verify this hypothesis, we first mask dimensions determined unessential by our algorithm, apply the masked word embeddings to a word sense disambiguation task (WSD), and compare its performance against the one achieved by the original embeddings.
Several KNN approaches are experimented to establish strong baselines for WSD.
Our results show that the masked word embeddings do not hurt the performance and can improve it by 3\%.
Our work can be used to conduct future research on the interpretability of contextualized embeddings.
% , thus proving our assumption that not all dimensions are equally important in representing word senses which could be 
\end{abstract}

\section{Introduction}
\label{sec:introduction}
Contextualized word embeddings have played an essential role in many NLP tasks. 
One could expect considerable performance boosts by simply substituting distributional word embeddings with Flair~\cite{15}, ELMo~\cite{14}, and BERT~\cite{16} embeddings.
% Enriched with sentential information, contextualized word embeddings tend to yield better performance. 
The unique thing about contextualized word embeddings is that different representations are generated for the same word type with different topical senses.
This work focuses on interpreting embedding representations for word senses. 
We propose an algorithm (Section~\ref{sec:approach}) that learns the dimension importance in representing sense information and then mask unessential dimensions that are deemed less meaningful in word sense representations to 0. 
The effectiveness of our approach is validated by a word sense disambiguation task (WSD) that aims to distinguish the correct senses of words under different contexts, as well as two intrinsic evaluations of embedding groups on the masked embeddings.

In addition to the final outputs of Flair, ELMo and BERT embeddings, hidden layer outputs from ELMo and BERT are also extracted and compared. 
Our results show that masking unessential dimensions of word embeddings does not impair the performance on WSD; moreover, discarding those dimensions can improve the performance up to 3\%, which suggests a new method for embedding distillation for more efficient neural network modeling.

\section{Related Work}
\label{sec:related-work}

\subsection{Word Embedding Interpretibility}
In the earlier work, \citet{59} suggest a variant of sparse matrix factorization, which generates highly interpretable word representations. Based on that work, \citet{70} introduce a method analyzing dimensions characterizing categories by linking concepts with types and comparing dimension values within concept groups with the average of dimension values within category groups. Works have also investigated ways to enrich embedding interpretability by modifying the training process of word embedding models~\cite{60,27}. Others make use of pre-trained embeddings and apply post-processing techniques to acquire embeddings with more interpretability. Past researches use matrix transformation methods on pre-trained embeddings~\cite{65, 66, 72}. \citet{65} utilizes canonical orthogonal transformations to map current embeddings to a new vector space where the vectors are more interpretable.

\noindent Similarly, \citet{66} proposes an approach that rotates pre-trained embedding by minimizing the complexity function, so that the dimensions after rotation become more interpretable. Another type of methods applies sparse encoding techniques on word embeddings and map them to sparse vectors~\cite{64,62}.

\subsection{Contextualized Word Embedding Models}

Three popular word embedding algorithms are used for our experiments with various dimensions: ELMo, Flair, and BERT. ELMo is a deep word-level bidirectional LSTM language model with character level convolution networks along with a final linear projection output layer~\cite{14}. Flair is a character-level bidirectional LSTM language model on sequences of characters~\cite{15}. BERT has an architecture of a multi-layer bidirectional transformer encoder~\cite{16}. 

\subsection{Word Sense Disambiguation (WSD)}
This work uses WSD as the evaluation for the proposed algorithm, which is the task of determining which sense a target word belongs to in a sentence. This work adopts a supervised approach that makes use of sense-annotated training data. The Most Frequent Sense (MFS) heuristic is the most common baseline, which selects the most frequent sense in the training data for the target word~\cite{wsd-base}. Depending on the evaluation dataset, the state-of-art in WSD varies. \citet{bilstm-att} utilize bi-LSTM networks with attention mechanism and a softmax layer. \citet{context2vec} and \citet{14} also adopt bi-LSTM networks with KNN classifiers. Later work incorporates word features such as gloss and POS information into memory networks~\cite{gas, supwsd}.

\section{Sense Weight Training (SWT)}
\label{sec:approach}

Given a large embedding dimension size, the hypothesis is that not every embedding dimension plays a role in representing a sense. Here we propose a new algorithm to determine the importance of dimensions. With word embedding groups classified by their senses annotated in the SemCor dataset~\cite{17}, the objective function in this algorithm is to maximize the average pair-wise cosine similarity in all sense groups. A weight matrix with the same size of the word embedding is initialized for each sense. Each dimension represents the importance of a specific dimension to that sense. 
% algorithm
\begin{algorithm}
\caption{Algorithm for Incremental Sense Weight Training}
\label{alg:training}
\begin{algorithmic}
\FOR{each sense group $SG$}
\STATE initialize weights $w$, learning rate $\gamma_{0}$, Adagrad weights matrix $gti$
\STATE initialize $S_{pre}$\\
$S_{pre} \leftarrow \sum\nolimits_{v_{i}, v_{j} \in SG, i\neq j} Cosine(v_{i}, v_{j})$
\FOR{each epoch $i$}
\IF {$i < n$}
\STATE randomly generate N numbers:\\
$D_1,\cdots, D_N$
\ELSE
\STATE generate N numbers based on policy: \\
$D_1,\cdots, D_N$
\ENDIF
\STATE $v_i [D_1,\cdots, D_N] \leftarrow 0$ for $v_i \in SG$
\STATE $S_{cur} \leftarrow \sum\nolimits_{v_{i}, v_{j} \in SG, i\neq j}Cosine(v_i, v_j)$ 
\STATE $grad = (S_{pre} -  S_{cur})* (mask-1) 
- \lambda * sign(w)$
\STATE $gti \mathrel{+}= grad^2$
\STATE $w \leftarrow \dfrac{w + grad * \gamma_{i}}{\epsilon+\sqrt{gti}} $
\ENDFOR
\ENDFOR
\end{algorithmic}
\end{algorithm}

\noindent During training, a mask matrix is generated and applied to the weight matrix. The gradient of the algorithm is defined to be the difference between the current similarity score and the previous similarity score multiplied by the masking matrix subtracted by one. The weight matrix is updated during training with the gradients and a learning rate. 
\par The mask matrix is the size of the weight matrix and has $N$ dimensions being zero and the rest being one. The generation of the mask matrix involves two phases. In the first phase, SWT randomly generates $N$ positions of zeros to ensure enough dimensions have been covered. After a certain number of epochs, the training enters the second phase where an  
\textbf{exploration-exploitation} policy is employed. The policy states that there is a chance of $\alpha$ to randomly generate $N$ numbers. For the remaining $1 - \alpha$ possibility, the generation of $N$ numbers depends on the weight matrix: the higher the value of dimension in the weight matrix, the lower probability of the number getting selected. Furthermore, $l_1$ regularization is applied for feature selection purpose, and AdaGrad \cite{Duchi:11} is used to encourage convergence. Pseudo-code for SWT is in Algorithm~\ref{alg:training}, where $n$ is the number of epochs for exploration, $\lambda$ the parameter for $l_1$ regularization and $\epsilon$ a small number to prevent zero denominators in AdaGrad. After the weights are learned, we set the value of embedding dimensions with low importance to zero and test if the rest dimensions are enough to represent the word sense group. 

\section{Experiments}
\label{sec:experiments}
% bert large graph
% bert base graph
\begin{figure*}[htpb!]
    \centering
    \includegraphics[width=.85\linewidth]{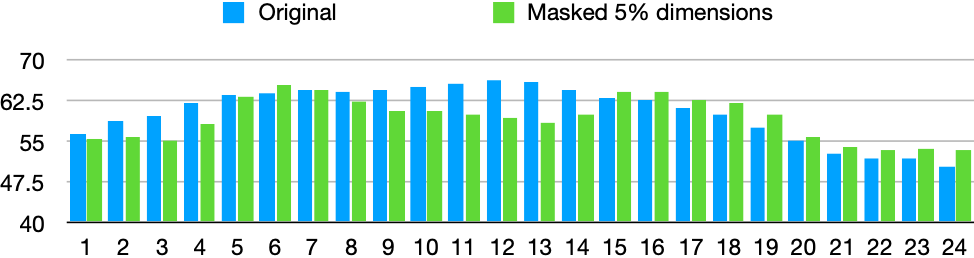}
    \caption{BERT-Large embeddings with 24 hidden layers. Certain layers such as the last 10 layers perform better if 5\% of the dimensions are masked.}
    \label{fig:bert-large}
\end{figure*}

% The task is to demonstrate the effectiveness of our algorithms to mask out the relatively useless dimensions, evaluated using WSD Evaluate Framework~\cite{wsd-framework} and F1 score. First, the best KNN method for the embeddings will be selected for all subsequent tasks. 
Firstly, all the experiments using the original word embeddings are run. Then, using the trained weight matrix from Section~\ref{sec:approach}, the same tests are run on masked embeddings again for comparison. 
% Section~\ref{ssec:datasets-embeddings} will talk about the training and evaluation datasets. Section~\ref{ssec:knn-methods} will discuss the baseline-finding approaches. Section~\ref{ssec: masked-embedding} will talk in details about how we mask the training and testing embeddings.
\subsection{Datasets and Word Embeddings}
\label{ssec:datasets-embeddings}
Our proposed baselines and algorithms are trained on SemCor~\cite{17} and evaluated on SenEval-2~\cite{seneval2}, SenEval-3~\cite{seneval3}, SemEval'07~\cite{semeval7}, SemEval'13~\cite{semeval13} and SemEval'15~\cite{semeval15}.
% \begin{itemize}
% 	\item SemCor is the large training corpora with a total of 226,040 sense annotations across 352 documents and the main corpus used for WSD tasks.
% 	\item SenEval-2 is originally annotated with WordNet 1.7 and consists of 2282 annotations for nouns, verbs, adverbs and adjectives. 
% 	\item SenEval-3 consists of three documents from editorial, news story and fiction domains and 1850 annotations for word senses.
% 	\item The SemEval-07 is the smallest and consists of three documents with only 455 noun and verb words annotated with WordNet 2.1.
% 	\item The SemEval-13 consists of 13 documents from various domains and only nouns are considered, which have 1644 sense annotations.
% 	\item The SemEval-15 is the most recent and consists of 1022 sense annotations with 4 document's from biomedical, mathematical/computing and social issues. 
% \end{itemize}
\par Pre-trained contextualized word embeddings are exclusively used and compared. Pre-trained ELMo, BERT and Flair models are tested. The models include ELMo's three models with dimension sizes of 256, 512 and 1,024 (all with 2-layer bi-LSTM), BERT's 2 models: BERT-base with a dimension size of 768 and 12 output layers;  BERT-Large with a dimension size of 1,024 and 24 layers, and Flair's single-layer bi-LSTM models with dimension sizes of 2,048 and 4,096.

% \footnote{https://allennlp.org/elmo}
% \footnote{https://github.com/zalandoresearch/flair}
% \footnote{https://github.com/google-research/bert}

\subsection{KNN Methods}
\label{ssec:knn-methods}
K-Nearest Neighbor (KNN) approach is adopted from both ELMo~\cite{14} and \textit{context2vec}~\cite{context2vec} to establish strong baseline approaches. % that offer the best results before our actual task.

% \paragraph{1-Classifier For All} This is the simplest adaptation of KNN using all the tokens from the training corpus and their annotated senses at once. During prediction, this approach simply finds the closest neighbor to the embedding of the target word. The total number of classifiers is 1. The results when $k=1$ is reported after different $k$ is experimented.
\paragraph{Sense-based KNN}
Adapted from ELMo~\cite{14} with $k=1$, words that have the same senses are clustered together, and the average of that cluster is used as the sense vector, which is then fitted using a one KNN classifier. Unseen words from the test corpus fall back using the first sense from WordNet~\cite{Fellbaum:98}.

\paragraph{Word-based KNN}
Following \textit{context2vec} \cite{15}, a cluster of each lemma occurrences in the training set is formed. Each word has a distinct classifier, which will assign labels based on $k$, where $k = \min(\#\ of\ occurrences,\ 5)$. Unseen words from test corpus fall back using the first sense from WordNet.

\subsection{Masked Embeddings}
\label{ssec: masked-embedding}
Each sense has a trained weight matrix from Section~\ref{sec:approach}. We process the weight matrix by experimenting four percentages (5\%, 10\%, 15\%, 20\%) to find the best threshold to mask out dimensions: the embedding dimensions with weight value ranked below such percentage are marked 0.  For evaluations, each target word tries all the masks of its appeared senses and selects the masking that produces the closest distance $d$, where $d$ is the sum of the distances from the masked word to its $k$-nearest neighbors. 
% baseline result
\begin{table}[htpb!]
\centering
\begin{tabular}{c|c|c|c|c}
& \bf WF & \bf W & \bf SF & \bf S \\ \hline\hline
ELMo*       & -    & -    & \bf69.0 & -                     \\
\textit{context2vec}* & - & \bf 69.6 & - & -               \\\hline
Flair      & 63.7    & 61.4    & 60.0      & 55.1                          \\
ELMo       & 63.8    & 61.5    & 63.9 & \bf59.0                     \\
BERT & \bf67.3 & 65.2 & 59.0      & 54.1                       \\
\end{tabular}
\caption{Results using 4 proposed KNN methods described in Section~\ref{ssec:knn-methods}. *: published results. WF: Word-based KNN with fall back using WordNet. W: Word-based. SF: Sense-based with fall back. S: Sense-based. Flair: forward and backward. ELMo: both biLM layers. BERT: concatenation of last 4 layers.}
\label{tbl: baseline}
\end{table}

\subsection{Results}
\label{ssec:results}
% masked result
\begin{table}[htpb!]
\centering
\begin{tabular}{c|c|c}
\bf Model & \bf Original & \bf Masked \\ \hline\hline
Flair-4096 & \textbf{63.7} & 62.1\\
Flair-2048 & 60.5 & \bf60.7\\\hline
BERT & \textbf{67.3} & 64.5\\\hline
ELMo & \textbf{63.8}  & 63.0\\
ELMo-256 & 61.5 & \bf62.3\\
ELMo-512 & 62.7 & \bf63.0\\
ELMo-1024 & 62.5 & \bf63.4
\end{tabular}
\caption{Results for the original and embeddings with 5\% dimensions masked. }
\label{tbl: masked}
\end{table}
\paragraph{Baselines} As shown in Table~\ref{tbl: baseline}, BERT-Large, and ELMo tend to achieve higher F1 using all four methods, and word-based KNN with fall back works better in general. Therefore, KNN-WF are used to conduct all subsequent tasks. 
% \begin{figure}[htpb!]
% \centering
% \includegraphics[width=\columnwidth]{./img/Flair.png}
% \caption{Flair and Flair-Fast unmasked and masked result.}
% \label{fig:flair}
% \end{figure}

% \paragraph{Hidden layers of BERT-Large} In Figure~\ref{fig:bert-large}, the results are when 5\% of the embeddings are masked based on the weight matrix we gained from Section~\ref{sec:approach}. Half the dimensions are improved if being masked using the 5\% threshold, especially the last 10 layers. Surprisingly, the last layer is boosted by 3\%. 

% \paragraph{Hidden layers of BERT-Base} In Figure~\ref{fig:bert-base}, the results are when 5\% of the embeddings are masked. The first 3 layers are improved by the masking, as opposed to the BERT-Large model.

% \paragraph{Hidden layers of ELMo} In Figure~\ref{fig:elmo}, both 5\% and 10\% masking are reported because for layer 1 and 7 10\% threshold works better than 5\% . For layer 3, 6 and 9, 5\% threshold surpassed the performance of the unmasked embeddings. 
\paragraph{Masked Results} The embeddings from all output layers of ELMo, BERT and Flair are evaluated. Table~\ref{tbl: masked} proves that for ELMo and Flair-2048, masking does not hurt the performance too much and for single layers, it even shows improvements. Figure~\ref{fig:bert-large} shows the results when 5\% of the embeddings are masked. Half the embeddings are improved if being masked using the 5\% threshold, especially the last 10 layer outputs. Surprisingly, the last layer output score is boosted by 3\%. 
% bert base
\begin{figure}[htpb!]
    \centering
    \includegraphics[width=\linewidth]{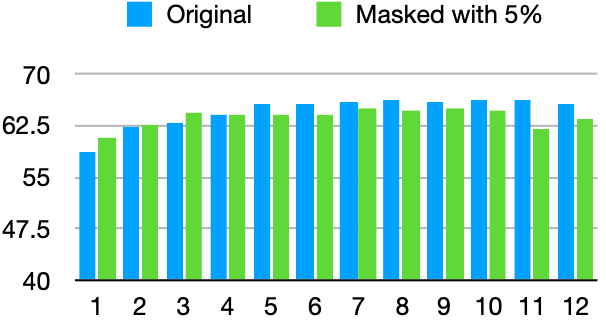}
    \caption{BERT-Base embeddings with 12 layers. The blue columns are original and the green ones are results after 5\% of the dimensions are masked.}
    \label{fig:bert-base}  
\end{figure}
% elmo graph

\begin{figure}[hptb!]
    \centering
    \includegraphics[width=\columnwidth]{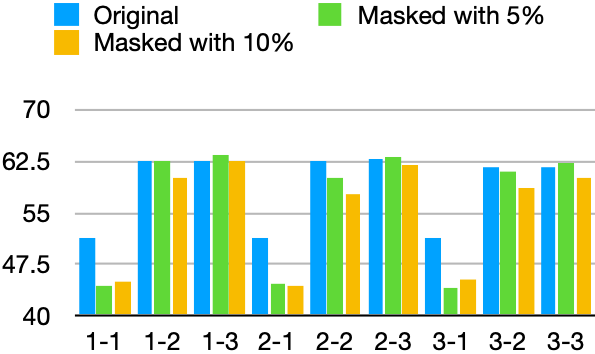}
    \caption{ELMo embeddings with 3 models and 3 layers each. 1: 1024. 2: 512. 3: 256. Blue columns are original embeddings, green columns are when 5\% of the dimensions are masked, and orange columns are when 10\% of dimensions are masked.}
    \label{fig:elmo}
\end{figure}

\noindent In Figure~\ref{fig:bert-base}, the first 3 layer outputs are improved by the masking with 5\% threshold. Why the deeper layer outputs are not improved requires further research. In Figure~\ref{fig:elmo}, both 5\% and 10\% masking are reported because for layer 1-1 and 3-1 10\% threshold works better than 5\%. For the last layer of each model, the 5\% threshold surpasses the performance of the original ones. 

ELMo performances vary more with different output layers, compared to BERT. BERT-Base output layers exhibit more stable performances compared to the BERT-Large model. Furthermore, an interesting pattern for ELMo is that masking out 5\% dimensions cause a more considerable performance drop for layers with worse original scores. One possible explanation is that embeddings from output layers closer to the input layer contain less insignificant dimensions. 

\noindent Experiments have also been done for the Flair models, which show similar results that the performances remain stable after 5\% dimensions of embeddings masked to zero, as shown in Table~\ref{ssec:results}.  
In summary, masking 5\% of the dimensions does not hurt the performance too much, and for half of them, masking helps improve the score by 3 percent at most. 10\% threshold sometimes outperforms the 5\% threshold in ELMo hidden layers.

\begin{figure*}[htpb!]
\begin{subfigure}{\columnwidth}
\centering
    \includegraphics[width=\linewidth]{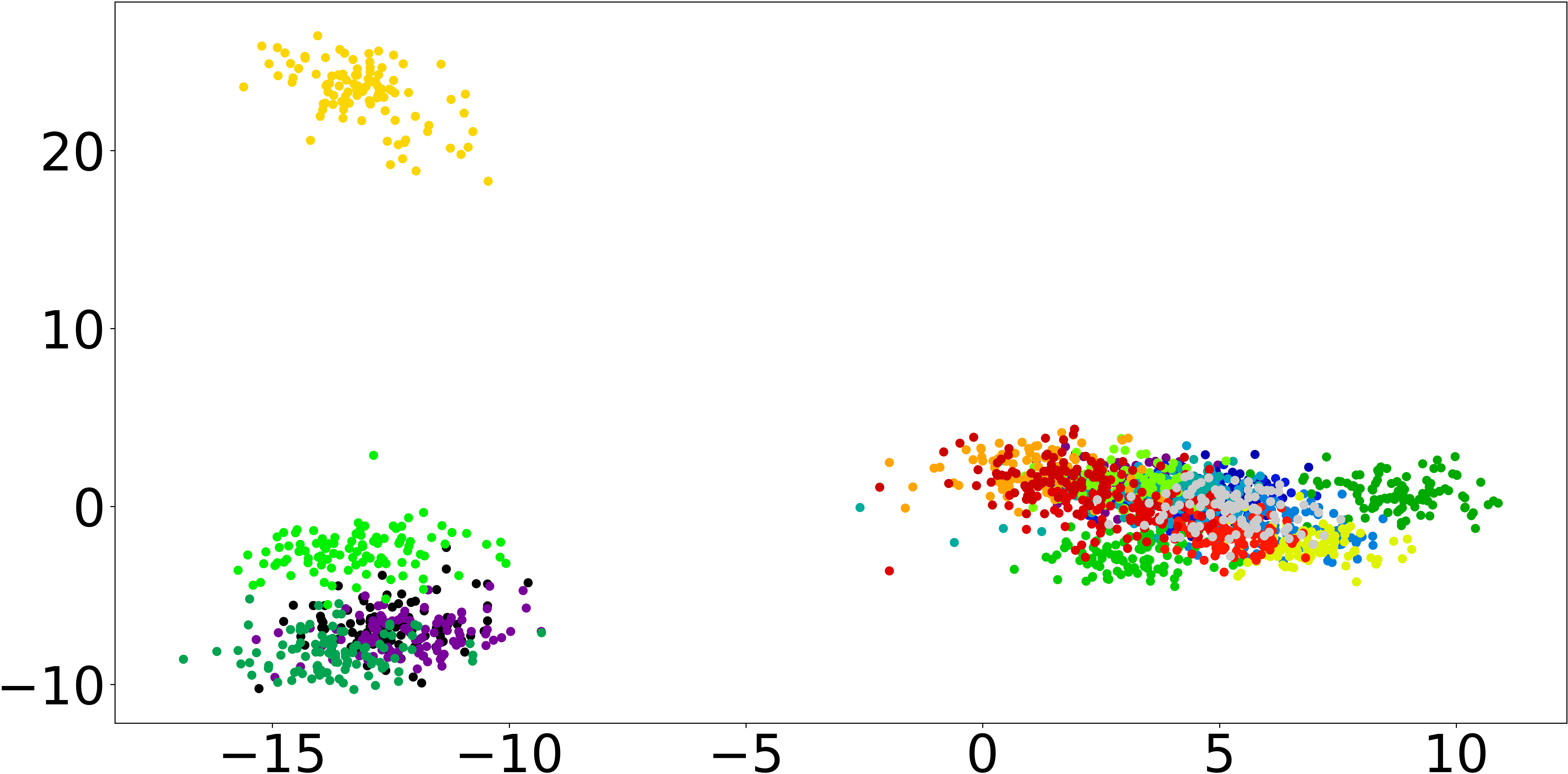}
    \caption{original}
    \label{fig:unmasked_cluster}
\end{subfigure}~~~~
\begin{subfigure}{\columnwidth}
    \centering
    \includegraphics[width=\linewidth]{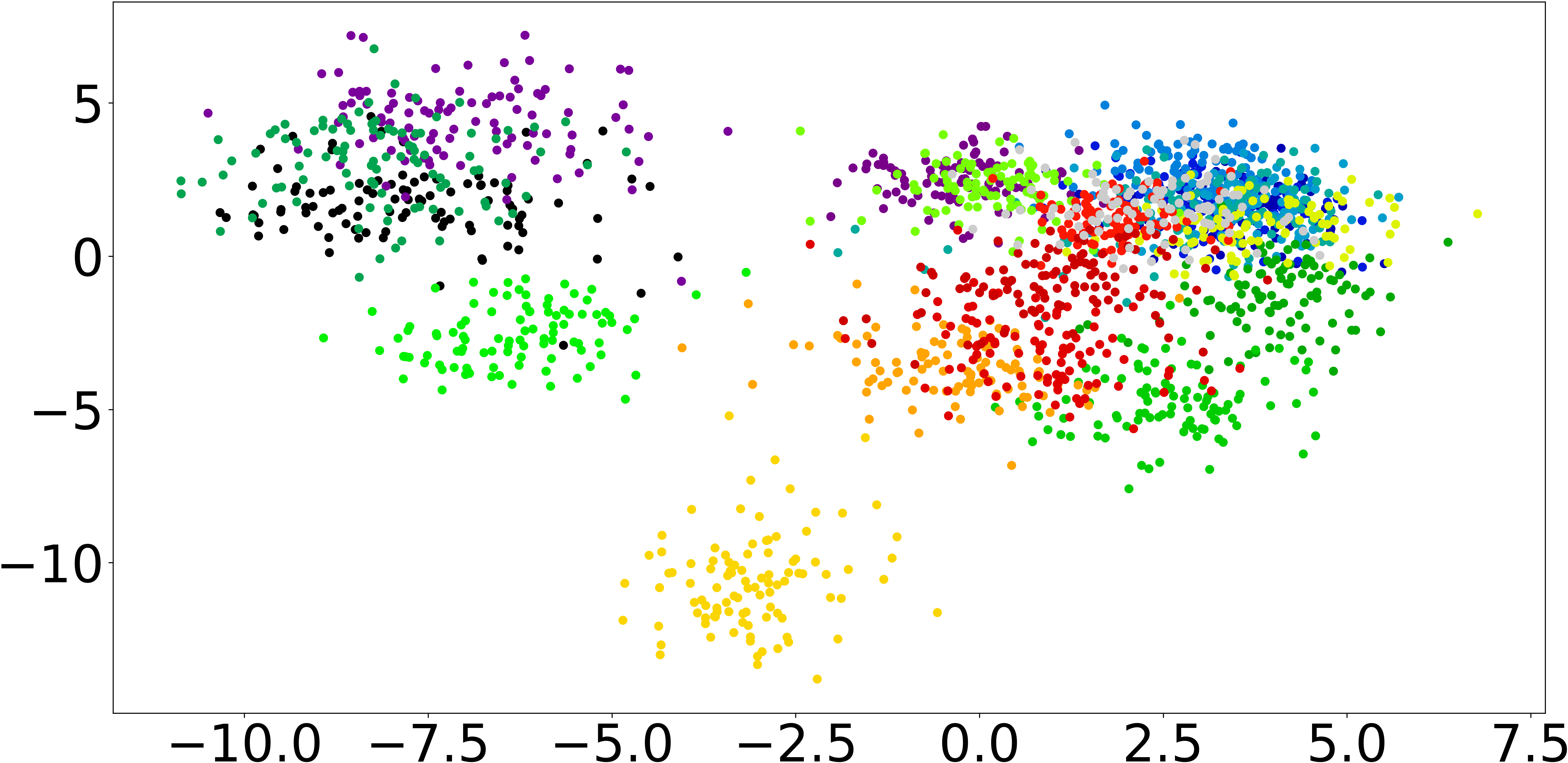}
    \caption{masked (threshold of 0.5)}
    \label{fig:masked_cluster}
\end{subfigure}
\caption{Graphs of 20 selected sense groups with 100 embeddings each for ELMo with a dimension size of 512 (third output layer). The projection of dimensions from 512 to 2 is done by Linear Discriminant Analysis.}
\end{figure*}
% analysis figure

% analysis
\subsection{Analysis}
\label{ssec:analysis}
Further analysis is made to investigate the number of negligible dimensions in word embeddings. Figure~\ref{fig:unmasked_cluster} shows a projected graph of selected sense groups, each with 100 embeddings from one ELMo model. Figure~\ref{fig:masked_cluster} demonstrates the same word embeddings with the dimensions masked to 0 if their corresponding weights are smaller than 0.5. The masked groups display a smaller with-in group distance and a greater separation of sense groups.

The Spearman's Rank-Order Correlation Coefficient $\rho$ between the pair-wise cosine similarity of sense vectors (average embedding of embedding groups classified by word senses) and the pair-wise path similarity scores between senses provided by WordNet~\cite{18} is evaluated for the original word embeddings and the masked embeddings whose group sizes are larger than 100.
Average pair-wise cosine similarity within sense groups is also calculated before and after. The table with all the test results is in the Appendix. Overall, the average cosine similarities within sense groups all increase after dimensions are masked out for all models, which proves that the dimension weights learn by our objective function. The correlation test shows no significant performance decrease (some even increase), which manifests that the masked dimensions do not contribute to the sense group relations. 

% analysis table
\begin{table}[htbp!]
\centering\small
\resizebox{\columnwidth}{!}{
\begin{tabular}{l|l|l|l|l}
\bf Model & \bf Dim   & \bf \bm{$N$}\textsubscript{masked} & \bf \bm{$\rho$}\textsubscript{original} & \bf \bm{$\rho$}\textsubscript{masked} \\ \hhline{=|=|=|=|=}
BERT  & 768   & 125        & 0.26814       & 0.26286        \\ 
BERT  & 1024  & 146        & 0.27423       & 0.26575         \\ \hline
ELMo  & 256    & 218        & 0.2852        & \textbf{0.3042}       \\ 
ELMo  & 512    & 281        & 0.29577       & \textbf{0.36943}    \\
ELMo  & 1024   & 608        & 0.28406       & \textbf{0.30675}    \\ 
 \hline
Flair & 2048  & 670        & 0.24891       & \textbf{0.28516}      \\ 
\end{tabular}}
\caption{Correlation coefficient test results $\rho$ for original and masked word embeddings with $N$\textsubscript{masked} (avg.\ number of dimensions masked out). The BERT embeddings are from the second output layer, and the ELMo and the Flair models from the last output layer.}
\label{tbl: analysis}
\end{table}

\noindent Table~\ref{tbl: analysis} contains the correlation test results and the number of dimensions masked out for BERT (the second to the last output layer), ELMo (the last output layer), and Flair. The number of dimensions masked is averaged throughout all sense groups. For ELMo and Flair, with the insignificant embedding dimensions masked out, the sense groups show a better correlation score. For the ELMo models, the number of embeddings that can be discarded increases with the distance of the output layer to the input layer. This result corresponds to ELMo's claim that the embeddings with output layers closer to the input layer are semantically richer~\cite{14}.

Another pattern is that the verb sense groups tend to have less number of dimensions getting masked out because verb sense groups have more possible forms of tokens belonging to the same sense group. A table with relevant examples can be found in the Appendix.  
We also attempted to mask out embedding dimensions with higher weights. In other words, we kept only the masked dimensions in the evaluations above, to examine what information is in the discarded dimensions. We ranked the cosine similarities between masked embedding pairs and picked out the 100 top most similar ones, which fails to output any patterns and points the future research direction in this domain.

\section{Conclusion}
This paper demonstrates a novel approach to interpret word embeddings. Mainly focusing on context-based word embedding’s ability to distinguish and learn relationships in word senses, we propose an algorithm for learning the importance of dimension weights in sense groups. After training the weights for word dimensions, the dimensions with less importance are masked out and tested using a word sense disambiguation task and two other evaluations. A conclusion can be drawn from the results that some dimensions do not contribute to the representation of sense groups and our algorithm can distinguish the importance of them. 
\label{sec:conclusion}

\clearpage

\bibliography{acl2019}
\bibliographystyle{acl_natbib}
\onecolumn
% \appendix
\section{Appendix}
\centering
\begin{table}[htpb!]
\centering
\scalebox{0.8}{
\begin{tabular}{l|l|l|l|l|l|l|l|l}
\hline
Model & Dim  & Out & Sense    & $N_{masked}$ & Sense      & $N_{masked}$ & Sense    & $N_{masked}$ \\ \hhline{=|=|=|=|=|=|=|=|=}
BERT  & 768  & -2  & ask.v.01 & 75           & three.s.01 & 208          & man.n.01 & 58           \\ 
BERT  & 1024 & -2  & ask.v.01 & 40           & three.s.01 & 211          & man.n.01 & 116          \\ \hline
ELMo  & 256  & 0   & ask.v.01 & 78           & three.s.01 & 182          & man.n.01 & 242          \\ 
ELMo  & 256  & 1   & ask.v.01 & 78           & three.s.01 & 99           & man.n.01 & 212          \\ 
ELMo  & 256  & 2   & ask.v.01 & 241          & three.s.01 & 243          & man.n.01 & 212                   \\ \hline
ELMo  & 512  & 0   & ask.v.01 & 103          & three.s.01 & 28           & man.n.01 & 295          \\ 
ELMo  & 512  & 1   & ask.v.01 & 144          & three.s.01 & 105          & man.n.01 & 156          \\ 
ELMo  & 512  & 2   & ask.v.01 & 334          & three.s.01 & 300          & man.n.01 & 311                    \\ \hline
ELMo  & 1024 & 0   & ask.v.01 & 174          & three.s.01 & 44           & man.n.01 & 331          \\ 
ELMo  & 1024 & 1   & ask.v.01 & 60           & three.s.01 & 106          & man.n.01 & 220          \\ 
ELMo  & 1024 & 2   & ask.v.01 & 568          & three.s.01 & 827          & man.n.01 & 708                   \\ \hline
Flair & 2048 & -1  & ask.v.01 & 193          & three.s.01 & 862          & man.n.01 & 1883         \\
\end{tabular}}
\caption{the embedding models with specific word embedding sense groups (Sense) and the embedding dimension numbers masked out in according groups ($N_{masked}$): ``ask.v.01" is a verb word sense with a meaning of ``inquire about"; ``three.s.01" is an adjective word sense with a meaning of ``being one more than two"; ``man.n.01" is a noun word sense with a meaning of ``an adult person who is male (as opposed to a woman)"}
\label{table5}
\end{table}

\begin{table}[htpb!]
  \centering
\scalebox{0.8}{
\begin{tabular}{l|l|l|l|l|l|l|l}
\hline
Model & Dim  & Out & $Dim_{masked}$ & $\rho_{unmasked}$ & $\rho_{masked}$ & $cos_{unmasked}$ & $cos_{masked}$ \\ \hhline{=|=|=|=|=|=|=|=}
BERT  & 768  & -2  & 125        & 0.26814       & 0.26286      & 0.4971   & 0.5187  \\ 
BERT  & 1024 & -2  & 146        & 0.27423       & 0.26575      & 0.5811   & 0.5983   \\ \hline
ELMo  & 256  & 0   & 95         & 0.12016       & \textbf{0.16017}      & 0.5959   & 0.6134  \\ 
ELMo  & 256  & 1   & 105        & 0.30903       & \textbf{0.37377}      & 0.5119   & 0.5787  \\
ELMo  & 256  & 2   & 218        & 0.2852        & \textbf{0.3042}       & 0.4507   & 0.6914  \\ 
ELMo  & 256  & av  & 199        & 0.26553       & \textbf{0.36945}      & 0.4932   & 0.6729  \\ \hline
ELMo  & 512  & 0   & 136        & 0.17058       & 0.17336      & 0.5957   & 0.6051  \\ 
ELMo  & 512  & 1   & 181        & 0.27967       & 0.25318      & 0.5414   & 0.5908   \\ 
ELMo  & 512  & 2   & 281        & 0.29577       & \textbf{0.36943}      & 0.4404   & 0.5346   \\
ELMo  & 512  & av  & 207        & 0.2949        & 0.30047      & 0.4930   & 0.5470  \\ \hline
ELMo  & 1024  & 0   & 179        & 0.18504       & 0.17263      & 0.5930   & 0.5945  \\
ELMo  & 1024  & 1   & 198        & 0.30897       & 0.30175      & 0.4783    & 0.4971   \\ 
ELMo  & 1024 & 2   & 608        & 0.28406       & \textbf{0.30675}      & 0.3927   & 0.4915    \\ 
ELMo  & 1024  & av  & 406        & 0.28331       & \textbf{0.27204}      & 0.4542   & 0.5086  \\ \hline
Flair & 2048 & -1  & 670        & 0.24891       & 0.28516      & 0.5560   & 0.6084  \\ 
\end{tabular}
}
\caption{Cosine similarity and correlation test results for unmasked and masked word embeddings: the embedding model (Model), dimension size (Dim), output layer (Out where av represents the average embedding of three output layers), the average number of dimensions that are masked to zero in embedding sense groups ($Dim_{masked}$), the correlation coefficient of original embedding sense group centers and sense relations ($\rho_{unmasked}$), the correlation coefficient of embedding sense group centers with dimensions masked to 0 ($\rho_{masked}$), the average within-group cosine similarity for the original embeddings ($cos_{unmasked}$) and the average within-group cosine similarity after the dimensions are masked out ($cos_{unmasked}$). Only sense groups with a group size bigger than 100 are considered in this case.}
\label{table4}
     
\end{table}

\begin{figure}[htpb!]
\centering
\includegraphics[width=.4\linewidth]{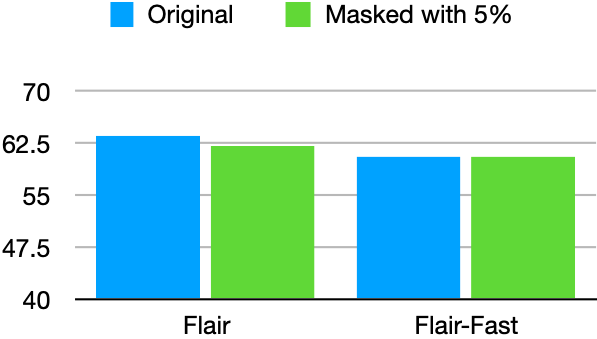}
\caption{Flair and Flair-Fast unmasked and masked result.}
\label{fig:flair}
\end{figure}

\twocolumn
\end{document}